\documentclass[fleqn,10pt]{wlscirep}

\usepackage[utf8]{inputenc}
\usepackage[T1]{fontenc}
\usepackage{multirow}
\usepackage{subcaption}
\usepackage{graphicx}
\usepackage{url}
\usepackage{amsmath}
\usepackage{mathtools}
\usepackage{tcolorbox}
\usepackage{booktabs}

\usepackage[pagewise]{lineno}
\usepackage{setspace}

\title{A Comparative Study on Patient Language across Therapeutic Domains for Effective Patient Voice Classification in Online Health Discussions}

\author[1,+]{Giorgos Lysandrou}
\author[1,+]{Roma English Owen}
\author[1]{Vanja Popovic}
\author[1]{Grant Le Brun}
\author[1]{Aryo Pradipta Gema}
\author[1]{Beatrice Alex}
\author[1,*]{Elizabeth A. L. Fairley}
\affil[1]{Talking Medicines Limited, SC447227, BECO Building, 58 Kingston Street, G5 8BP, Glasgow, Scotland, UK}

\affil[*]{Corresponding author: elizabeth@talkingmedicines.com}

\affil[+]{These authors contributed equally to this work}


\begin{abstract}

There exists an invisible barrier between healthcare professionals' perception of a patient's clinical experience and the reality.
This barrier may be induced by the environment that hinders patients from sharing their experiences openly with healthcare professionals.
As patients are observed to discuss and exchange knowledge more candidly on social media, valuable insights can be leveraged from these platforms.
However, the abundance of non-patient posts on social media necessitates filtering out such irrelevant content to distinguish the genuine voices of patients, a task we refer to as patient voice classification.
In this study, we analyse the importance of linguistic characteristics in accurately classifying patient voices.
Our findings underscore the essential role of linguistic and statistical text similarity analysis in identifying common patterns among patient groups.
These results allude to even starker differences in the way patients express themselves at a disease level and across various therapeutic domains.
Additionally, we fine-tuned a pre-trained Language Model on the combined datasets with similar linguistic patterns, resulting in a highly accurate automatic patient voice classification.
Being the pioneering study on the topic, our focus on extracting authentic patient experiences from social media stands as a crucial step towards advancing healthcare standards and fostering a patient-centric approach.

\end{abstract}

\begin{document}

\flushbottom
\maketitle
%
%
\thispagestyle{empty}


\section*{Introduction}


There is a critical need for global healthcare systems to provide better treatments for patients.
A substantial aspect of the shortcomings in healthcare systems worldwide can be traced back to the generalized nature of services and medications provided.
More personalized care (i.e. offering patients the right drugs, at the right time, in the right dose and formulation) holds the key to significantly improving outcomes~\cite{grissinger2010five}.

To achieve this, it is important to transition from a primarily professional-centric approach to a patient-centric approach, where we put more emphasis on listening to patient experiences.

However, capturing genuine patient experiences proves to be challenging.
Conventional methods, such as relying on the conclusions of healthcare professionals (HCPs) after their interactions with patients, often fall short of providing a comprehensive understanding~\cite{eland1999attitudinal}.
Patients' interactions with HCPs may be highly influenced by environmental factors, leading to less descriptive conversations and an incomplete portrayal of patients' experiences~\cite{davison2000talks, leonard2017exploring}.
Another conventional method is through patient focus groups, where the patients' interactions are recorded to collect their perspectives on healthcare-related topics, such as their experience using certain drugs.
Unfortunately, patient focus groups risk introducing bias by drawing feedback primarily from specific socioeconomic segments, potentially excluding a representative sample of the entire patient population.

Recognizing the limitations of traditional approaches, this study advocates for the exploration of an alternative medium, that is social media.
Patients are observed to share personal health-related details, facilitated by the distance and anonymity provided by online platforms~\cite{colineau2010talking,antheunis2013patients}.
The absence of HCPs in this space allows patients to express themselves more candidly and descriptively, providing a unique perspective to understand their experiences.
Compared to other conventional non-physical sources such as market research and pharmaceutical representatives' feedback, social media offers a more accessible and diverse pool of patient information, reducing potential biases.

The increase in studies focusing on analyzing social media data for patient experience information in recent years attests to the growing recognition of its potential~\cite{wessel2024using}.
Advances in data capture and analysis, coupled with the rise in relevant social media data, notably accelerated by the COVID-19 pandemic, have triggered this trend.
Platforms like Reddit\footnote{\url{https://www.reddit.com/}}, with international user bases organized around specific health topics (e.g. \texttt{r/eczema}\footnote{\url{https://www.reddit.com/r/eczema/}}), serve as hubs for sharing experiences and knowledge, illustrating the global reach and depth of such data.
Compellingly, research indicates that the quality of self-reported patient experiences on these platforms is often comparable to that provided by healthcare professionals~\cite{blenkinsopp2007patient, van2008coping}, suggesting the validity and richness of patient narratives found in social media discourse.

Amidst this backdrop, we hypothesize that patient experience information cannot be uniformly treated across therapeutic domains and data sources due to inherent differences in how individuals articulate their experiences.
To test this hypothesis, the study adopts a multi-faceted approach, collecting data from social media (Reddit) and message boards (SocialGist) in specific therapeutic domains (i.e. cardiovascular, oncology, immunology, and neurology (COIN)).
Initial top-level analysis shows distinct subsets of data, requiring a deeper exploration of the linguistic nuances within each domain.

The preceding work by Alex et al. (2021)~\cite{alex2021classifying} focuses on the identification of patient voices in social media posts.
Their classifier model, trained on English language data from Reddit and Twitter, demonstrated the importance of tailoring models to specific therapeutic domains, showcasing a minor performance increase.
Building upon this foundation, our study aims to delve deeper into linguistic differences within therapeutic domains, contributing to a more nuanced understanding and optimal classification of patient voices.

In summary, our contributions are as follows:

\begin{itemize}
    \item Identifying an appropriate linguistic and text similarity analysis to guide and explain the patient voice classification,
    \item Understanding the commonalities and differences between the ways patients express themselves, about different health conditions, depending on therapeutic domain and data source,
    \item Determining the optimal machine learning (ML) classification methodology to classify posts belonging to patients, amongst different therapeutic domains and data sources.
\end{itemize}

\section*{Related Work}

\subsection*{Natural Language Processing in Healthcare}

The digitization of healthcare has experienced a substantial upswing in recent years.
Global events, such as the COVID-19 pandemic, have triggered a surge in patient engagement within online platforms.
This transition is complemented by the rapid advancements in Artificial Intelligence (AI), which have shown remarkable efficacy in enhancing healthcare data analysis.
In this context, the employment of Natural Language Processing (NLP) and Deep Learning (DL) techniques has become increasingly pivotal in interpreting both structured and unstructured healthcare data~\cite{lee2020biobert, alsentzer2019publicly, lehman2023clinical, gema2023parameter}.


A survey study by Hudaa et al. (2019)~\cite{hudaa2019natural} reported the instrumental role of NLP in extracting information from unstructured healthcare data, through methods such as document classification and feature extraction.
Their research highlights the advent of Large Language Models (LLMs) as a breakthrough in understanding natural language, positing substantial improvements in patient-healthcare provider interactions.

Similarly, Lavanya et al. (2021)~\cite{lavanya2021deep} investigated the utilization of DL techniques for classifying healthcare-related texts within social media platforms.
Given the rapid emergence of social media as a predominant platform for healthcare discourse, their findings stress the critical need for optimized information extraction methodologies.

\subsection*{Patient Voice Detection}

The study of patient voices within the social media sphere further exemplifies the utility of NLP in healthcare.
Pattisapu et al. (2017)~\cite{pattisapu2017medical} demonstrated an approach to discern medical personas through social media posts, utilizing pre-trained word embeddings (e.g. Word2Vec~\cite{mikolov2013distributed}) for superior information extraction.
This analysis not only aids in understanding patient perspectives but also serves pivotal roles in drug marketing and safety monitoring.

Dai et al. (2017)~\cite{dai2017social} investigated the clustering of social media post embeddings as an advanced alternative to traditional classification methods. Their unsupervised method, relying on no labelled training data, showcased a commendable classification accuracy, heralding a new era of understanding social media discourse in healthcare.

Alex et al. (2021)~\cite{alex2021classifying} provided an extensive review of patient voice detection in their study~\cite{jiang2016construction, sewalk2018using, zhu2020identifying, lu2020user, okon2020natural, meeking2020patients}.
Their work amplifies the discourse on patient voice detection, setting the stage for future innovations in this field.

\subsection*{Linguistic Analysis of Patient Language}

The linguistic analysis of patient language offers another dimension of insight into patient experiences and concerns.
Lu et al. (2013)~\cite{lu2013health} performed topic analysis within online patient communities, uncovering prevalent discussions on symptoms, drugs, and procedures.
Their findings illustrated the diversity of patient concerns, varying significantly across different disease-specific communities.

Dreisbach et al. (2019)~\cite{dreisbach2019systematic} provided a review of NLP applications in extracting clinical symptoms from patient-authored texts across various platforms, such as Twitter and online community forums.

\section*{Data Acquisition and Annotation}

\subsection*{Data Sources and Collection Methodology}

For the experiments detailed herein, data was systematically collected from two principal online platforms: Reddit\footnote{\url{https://www.reddit.com/}} and SocialGist\footnote{\url{https://socialgist.com/}}.
We selected these platforms for their extensive user-generated content on health-related topics.
Reddit, known for its user-created communities called \textit{subreddits}, provided a diverse range of discussions across various health conditions including cardiovascular, oncology, immunology, and neurology.
SocialGist, serving as a data aggregator, offered access to a wide array of message board posts from multiple community websites focusing on similar health domains.

We utilized the Pushshift Reddit API\footnote{\url{https://github.com/pushshift/api}} to retrieve a comprehensive list of historical and current posts from targeted subreddits.
Similarly, the SocialGist API facilitated the collection of message board posts.
A meticulously curated list of search terms, related to specific drugs and therapies within the aforementioned therapeutic areas, guided the data retrieval process.
This approach is supported by literature indicating that carefully selected search terms can yield high levels of precision and recall in data collection efforts~\cite{llewellyn2015extracting}.

Duplicate entries were identified and removed based on the text body and unique identifiers.
The final dataset comprised 14,693 posts, with an almost equal distribution between Reddit (7,211 posts) and SocialGist (7,482 posts).
A detailed breakdown of the data volumes by source and therapeutic domain is available in Supplementary Material 1.

\subsection*{Manual Annotation}

Subsequent to data collection, the posts underwent a manual annotation process.
Utilizing Doccano\footnote{\url{https://doccano.github.io/doccano/}}, an open-source annotation tool, a team of trained annotators applied document-level labels to each post.
These labels distinguished between \textit{``Patient Voice''} and \textit{``Not Relevant''}.
\textit{``Patient Voice''} denotes first-hand experiences of patients, while \textit{``Not Relevant''} denotes all other content types, including healthcare professional insights, news articles, etc.
To ensure reliability, the annotators adhered to detailed guidelines continuously refined throughout the project. 
Examples of each label are as follows:

\begin{itemize}
    \item \textbf{Patient Voice}: \textit{"I'm taking MTX and imraldi at the moment, so far so good." }
    \item \textbf{Not Relevant}: \textit{"MHRA due to approve new RA drug.", "One of my old patients used to take 10mg eliquis instead of 5mg. His heart rate was..." }
\end{itemize}

\begin{figure}
    \centering
    \includegraphics[width=\textwidth]{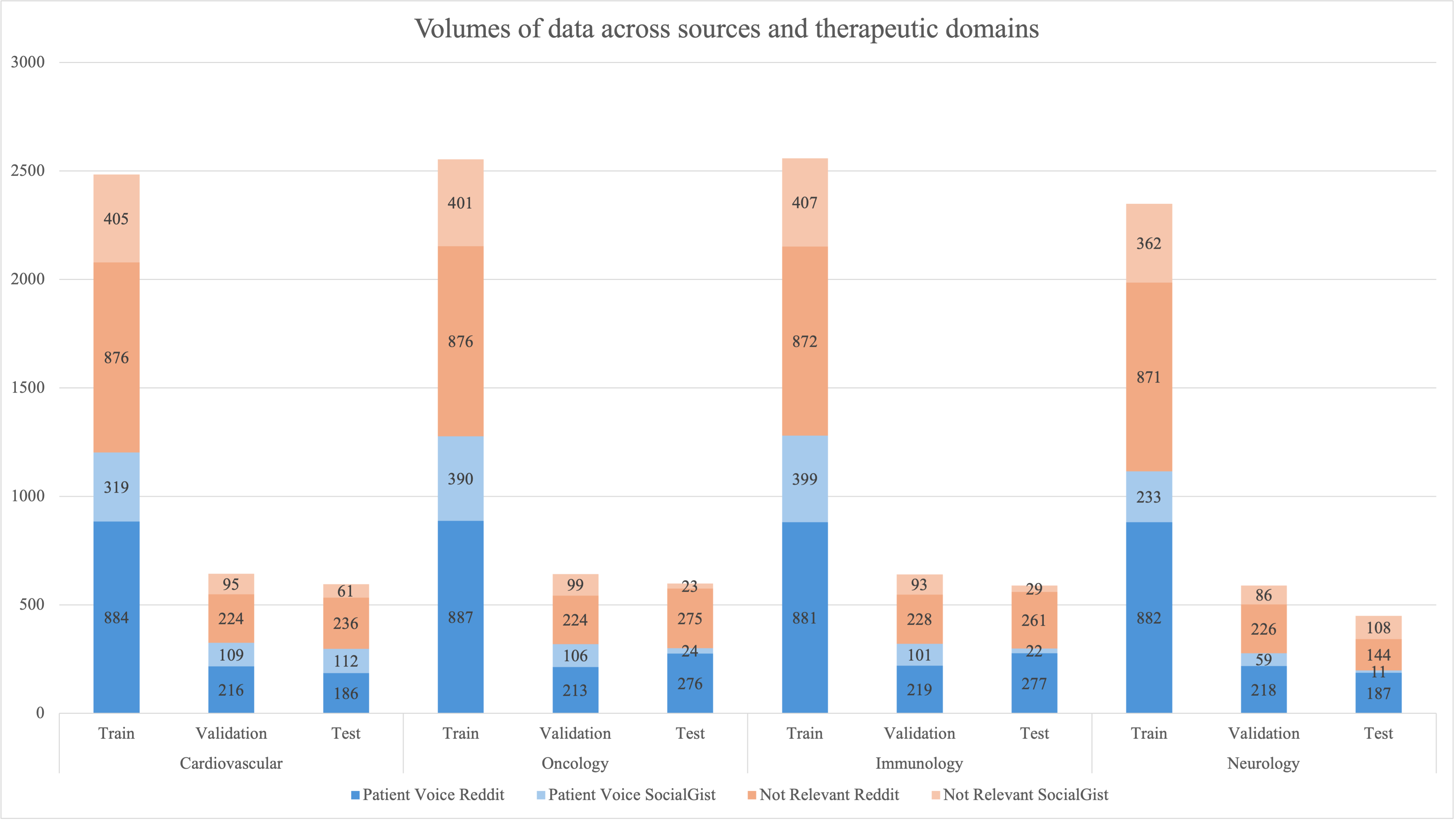}
    \caption{Experiment data volumes across data sources and splits, grouped by the therapeutic domains. The labels are ``Patient Voice'' and ``Not Relevant''.}
    \label{fig:experiment_data_volumes}
\end{figure}

After a single annotation phase, the dataset was partitioned into training (80\%) and validation (20\%) subsets.
This division was executed post-randomization with a reproducible random seed to mitigate bias while maintaining consistent label distribution across splits.
We conducted an additional annotation phase to collect a holdout test set which is equal in size to the validation subset for classifier evaluation. The resulting train validate and test ratios are 66\% train, 17\% validation and 17\% test.
We also ensured that no validation or test data were leaked into the training data.
\autoref{fig:experiment_data_volumes} details the distribution of label counts across all domains and both data sources of the data.

For each therapeutic domain, we merged data from Reddit and SocialGist (e.g., all cardiovascular posts from both sources form a single cardiovascular dataset).
Furthermore, we created a combined dataset comprising posts from all domains and platforms, denoted as ``All'' data.
The distribution statistics for these consolidated datasets are detailed in~\autoref{tab:experiment_combined_data_volumes}.

\begin{table}[t]
\centering
\caption{Experiment combined data volumes across datasets with combined data sources, therapeutic domains, train, validation and test splits, for the classes of Patient Voice and Not Relevant social media posts.}
\begin{tabular}{lcccccc}
\toprule
\multirow{2}{*}{\textbf{Combined Datasets}} & \multicolumn{2}{c}{\textbf{Train}} & \multicolumn{2}{c}{\textbf{Validation}} & \multicolumn{2}{c}{\textbf{Test}} \\
 & \textbf{Patient Voice} & \textbf{Not Relevant} & \textbf{Patient Voice} & \textbf{Not Relevant} & \textbf{Patient Voice} & \textbf{Not Relevant} \\
\midrule
Cardiovascular  & 1760 & 724 & 440 & 204 & 422 & 173 \\
Oncology        & 1763 & 791 & 437 & 205 & 551 & 47 \\
Immunology      & 1753 & 806 & 447 & 194 & 538 & 51 \\
Neurology       & 1753 & 595 & 544 & 145 & 331 & 119 \\
\midrule
All             & 7029 & 2916 & 1768 & 748 & 1842 & 390 \\
\bottomrule
\end{tabular}
\label{tab:experiment_combined_data_volumes}
\end{table}
\subsection*{Inter-Annotator Agreement}

We calculated the Inter-Annotator Agreement (IAA) scores to evaluate the consistency among annotators and the effectiveness of our annotation guidelines.
These scores are also important as they indicate the maximum performance our AI model could achieve should it succeed in modeling human classification accuracy.
We involved 12 annotators to label 2,388 posts selected at random from all four therapeutic domains.
For each annotator pair, we computed standard metrics, precision, recall, and F1 score, both weighted and macro-averaged, alongside Cohen's Kappa score.
Cohen's Kappa score accounts for chance agreement, offering a more robust assessment than the other three scores alone.
With an average Cohen's Kappa score of 0.773, we observed significant agreement among annotators.
This finding affirms the annotation guidelines' clarity and the dataset's reliability.



\section*{Methods}

\begin{figure}[t]
    \centering
    \includegraphics[width=\textwidth]{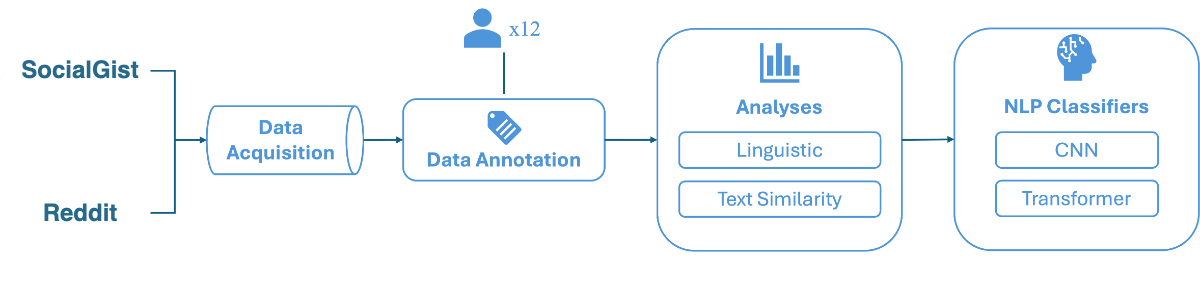}
    \caption{The pipeline for classifying patient voices from online health communities. Starting with data acquisition from Reddit and SocialGist, the methodology encompasses a sequential process from annotating the data, partitioning into train-validation-test splits, comprehensive linguistic and textual analyses, as well as the training of NLP classifier models.}
    \label{fig:schema}
\end{figure}

The workflow of our study is visualized in~\autoref{fig:schema}.
We start with the data collection as stated in the previous chapter, followed by annotation and data partitioning.
This fuels our subsequent qualitative linguistic and statistical text analyses.
Our workflow is concluded with the development of NLP classifiers trained to identify patient voices.

\subsection*{Qualitative Linguistic Analysis}

Our annotation team conducted a qualitative analysis of randomly selected posts from each therapeutic domain and data source-specific dataset to gather apparent differences in the language that patients use in describing their experiences.

\subsection*{Statistical Text Similarity Analysis}

We measure the linguistic similarity between datasets by calculating cosine similarities of datasets' vector representations.
We leverage Term-Frequency Inverse-Document-Frequency (TF-IDF)~\cite{sparck1972statistical} to calculate a representation for each word which indicates its value of significance based on the number of its occurrences within one dataset and across multiple datasets:
\begin{equation}
    TFIDF(t, d) = \left( \frac{f_{t, d}}{\sum_{t' \in d} f_{t', d}} \right) \times \log \left( \frac{|D|}{|\{d \in D : t \in d\}|} \right)
\end{equation}
where \(f_{t, d}\) is the frequency of term \(t\) in document \(d\), \(\sum_{t' \in d} f_{t', d}\) is the total number of terms in document \(d\), \(|D|\) is the total number of documents in the corpus, and \(|\{d \in D : t \in d\}|\) is the number of documents where the term \(t\) appears.

Subsequently, we can generate a dataset's vector representation by aggregating the TF-IDF vectors of all words inside a dataset, effectively summarizing the dataset's lexical characteristics.
Utilizing these vector representations, we conduct pairwise comparisons of datasets from various data sources and therapeutic domains by calculating the cosine similarities between pairs of datasets:
\begin{equation}
    cos(D_A, D_B) = \frac{D_A \cdot D_B}{||D_A|| * ||D_B||}
\end{equation}
where $D_A$ and $D_B$ denote the vector representation of dataset A and dataset B, respectively.

\subsection*{Text classifiers}

In our study, we focus on identifying patient voices within online posts using advanced text classification techniques.
We employ two model architectures:

\begin{itemize}
    \item \textbf{Convolutional Neural Network (CNN) Text Classifier}: We use spaCy's~\cite{honnibal2020spacy} small CNN \texttt{en\_core\_web\_sm} as the baseline of the experiment.
    This model uses mean pooling and attention mechanisms within its CNN architecture.
    We chose this architecture for its balance of efficiency and accuracy.
    We refer to this model as \textit{``CNN''} classifier in later sections.
    \item \textbf{Transformer-Based Classifier (RoBERTa)}: For a more computationally intensive and accurate solution, we leverage spaCy's transformer model \texttt{en\_core\_web\_trf}. 
    It uses a RoBERTa~\cite{liu2019roberta} base model which has been pre-trained on a large general-domain text corpus, providing contextually rich word representations.
    We refer to this model as \textit{``Transformer''} classifier in subsequent sections.
\end{itemize}

We employ two different model architectures in our study: a CNN and a transformer-based model, specifically RoBERTa. Both models utilize a bag-of-words approach for document representation, where each word in a text is represented by a vector created by the language models. These word vectors are context-dependent, as they capture the contextual meaning of words within the sentence they appear in. The document representation is then constructed by combining the word representation vectors for all the words in the text.

We use spaCy's TextCategorizer module for the modelling needs of this work. It supports multiple language model architectures, which use a bag-of-words approach to classify text. We train the models on our training data, seen in Figure~\ref{fig:experiment_data_volumes} and Table~\ref{tab:experiment_combined_data_volumes}, such that they are fine-tuned for our specific task of identifying patient voice amongst online posts.

\subsection*{Evaluation metrics}

We report the standard metrics of precision, recall and F1 scores for each label type, weighted and macro averaged F1 scores across all label types. The classifier models are evaluated on the test datasets.

\subsection*{Statistical Validation}

To assess the statistical significance of the performance differences between our classifiers, we employed McNemar's test ~\cite{dietterich1998approximate}. This test evaluates whether the differences in performance are due to random variation or represent true differences. The test takes as input the predictions of two classifiers on the same test dataset and calculates the likelihood of these differences occurring by chance.

\section*{Results \& Discussions}

\subsection*{Qualitative Linguistic Analysis}


Manual analysis and review of data from each therapeutic domain and data source specific dataset, shows differences in the language patients use to describe their experiences. Quantitative analysis between the two data sources, showed that overall, SocialGist patients describe their experiences using more words (longer posts), compared to Reddit patients, while also using a richer vocabulary. This difference is reflected in the average total unique words to total words per post ratio.

The methodology for performing a qualitative linguistic analysis on the experiment data involves several steps. Firstly, a random selection is made throughout the dataset, to mitigate selection bias, choosing 10 posts from each therapy area and data source, adding up to 20 posts per therapy area. These posts are then manually examined to identify common characteristics within the language used by patients to express their experiences, supplementing from past experience through exposure to the data. Subsequently, a single post or extract from a post that best exemplifies these identified characteristics is selected. The selected extract serves as a representative sample of the therapeutic domain, showcasing the unique language variations within each domain. The following are the examples illustrating these therapy area specific characteristics:

\begin{itemize}
    \item \textbf{Cardiovascular:} \textit{``I had a blood test for my D levels. They were scarily low the year of my heart attack (2016). The doctor put me on 1000mg a day of D3.''}
    \item \textbf{Oncology:} \textit{``I'm in a similar situation to you! I'm 32, was diagnosed in April, I have HER2+ invasive Stage 2B grade 2 with lymph node involvement.''}
    \item \textbf{Immunology:} \textit{``I now take Cosentyx. However, after 3 months I have noticed what looks either like psoriasis on soles of my feet or could be athlete’s foot. After 15 weeks, I now have the same peeling all over palms, between fingers, and backs of fingers.''}
    \item \textbf{Neurology:} \textit{``I have taken quetiapine, abilify and olanzapine. Olanzapine didn’t help at all and I always feel a bit nervous and anxious on the quetiapine.''}
\end{itemize}
%

Patients posting on cardiovascular-related conditions often mention detailed references to medication dosages, side effects, lifestyle factors, as well as significant health events, such as \textit{``heart attack''} or \textit{``stroke''}.
This pattern suggests that cardiovascular patients primarily reflect on the contrast between their pre-diagnosis lifestyle (e.g. weight characteristics) and their condition post-diagnosis.
The nature of cardiovascular symptoms, which are typically less visible without medical intervention, possibly contributes to this focus.
On the rarer occasion that patients are made aware of their symptoms ahead of crisis point, it is still difficult to notice these symptoms themselves to the same extent an immunology patient with itchy skin could.
High blood pressure for instance would be almost impossible to detect for oneself without the aid of a wearable device.
When medicines are discussed in this cardiovascular context, the patient conversation often focuses on the side effects of these medicines, that can be detected by a patient, rather than the disease-related symptoms.

In contrast, oncology discussions were characterized by a high degree of specificity, with patients frequently discussing genetic markers, clinical trials, and the outcomes of diagnostic tests.
Specific terms such as \textit{``HER2+''}, \textit{``ER-''}, and \textit{``Stage IV''} are commonly found in oncology posts.
This may indicate a reliance on the information given by the health care professionals for understanding and communicating their condition.
In our experiment, oncology patients' posts contained the most concise and specific language.

Conversations among immunology patients were noticeably detailed, reflecting the visible and impactful nature of their symptoms in their daily lives.
These discussions often included vivid descriptions of physical symptoms, highlighting the significant effect of immunological conditions on patient well-being.
As shown in the example, patients tend to share detailed experiences such as body parts that are affected and exact sensations that they experienced.

Neurological condition discussions were predominantly centered around the subjective experience of living with the condition, with a strong emphasis on emotional well-being and the effects of various treatments.

\subsection*{Statistical Text Similarity Analysis}

To complement our qualitative insights, we conduct a statistical text similarity analysis to quantitatively assess lexical similarities across datasets.
Utilizing cosine similarity measures derived from TF-IDF vectors, we examine the linguistic commonalities between datasets spanning different data sources and therapeutic domains.
This analysis focuses on the stemmed corpus of each dataset, with stopwords removed to highlight semantic parallels more effectively.

\begin{figure}[t]
\centering
\includegraphics[scale=0.45]{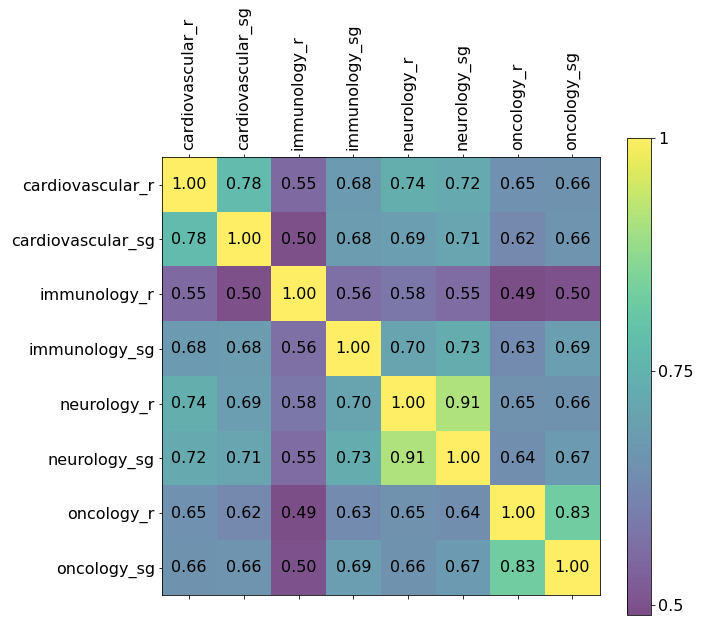}
\caption{Pairwise comparison matrix of cosine similarity values between all data source and therapeutic domain-specific subsets of the data, after TF-IDF analysis.}
\label{fig:tf_idf}
\end{figure}

The pairwise similarity is illustrated in~\autoref{fig:tf_idf}.
A value of 1.0, depicted in bright yellow, indicates a high degree of lexical similarity.
In contrast, darker blue shades denote lower similarities, suggesting diverse vocabularies between datasets.
Through this methodology, we consider three categories of similarity: low (0.45-0.60), medium (0.60-0.75), and high (above 0.75).
18 dataset pairs fell within the medium similarity range, seven pairs displaying low similarity and three pairs exhibiting significant similarity.

Notably, the analysis showed that patient posts within the same therapeutic domain, but across different data sources, tend to share a substantial amount of linguistic commonality.
This trend was particularly evident among cardiovascular, neurology, and oncology discussions across Reddit and SocialGist platforms.
Conversely, immunology patient discourse displayed marked language differences between these two sources, with Reddit's immunology dataset (denoted as \texttt{immunology\_r}) registering the lowest cosine similarity scores in comparison to all other datasets, including those from SocialGist (denoted as \texttt{immunology\_sg}).

To understand these findings, we identified the top 20 words with the highest TF-IDF scores within each dataset as shown in~\autoref{tab:tfidf_ranked}.
This analysis confirmed that Reddit's immunology discussions are characterized by a distinctive lexicon, with 13 of the top 20 TF-IDF terms being unique to this dataset.
Similarly, SocialGist's immunology content featured 10 unique terms among its top 20, frequently referencing specific medications such as \textit{``Humira''}, \textit{``Ocrevus''}, and \textit{``Entyvio''}.
This contrasts with other therapeutic areas, where the top TF-IDF terms often encompassed temporal references (e.g., ``days'', ``weeks'', ``years'') and medical terminology (e.g., ``doctor'', ``meds'', ``treatment''), alongside expressions related to the patient experience (e.g., ``pain'', ``help'', ``effects'').

\begin{table}[t]
\centering
\caption{Top 20 words per dataset, ranked by their TF-IDF score in descending order, with \textbf{bold} denoting words common in more than one dataset's top 20 words list.}
\begin{tabular}{llllllll}
\toprule
\multicolumn{2}{c}{\textbf{Cardiovascular}} & \multicolumn{2}{c}{\textbf{Oncology}}         & \multicolumn{2}{c}{\textbf{Immunology}}  & \multicolumn{2}{c}{\textbf{Neurology}}    \\
\textbf{Reddit}          & \textbf{SocialGist}       & \textbf{Reddit}             & \textbf{SocialGist}         & \textbf{Reddit}             & \textbf{SocialGist}         & \textbf{Reddit}          & \textbf{SocialGist}       \\
\midrule
stroke          & \textbf{blood}   & \textbf{arimidex}  & \textbf{chemo}     & eczema             & \textbf{years}     & \textbf{feel}   & \textbf{abilify} \\
\textbf{heart}  & eliquis          & week               & \textbf{herceptin} & skin               & humira             & \textbf{taking} & \textbf{zyprexa} \\
cholesterol     & \textbf{heart}   & test               & \textbf{treatment} & \textbf{psoriasis} & entyvio            & \textbf{really} & \textbf{time}    \\
\textbf{know}   & \textbf{day}     & \textbf{cancer}    & \textbf{years}     & face               & \textbf{time}            & zoloft          & \textbf{taking}  \\
\textbf{blood}  & \textbf{years}   & \textbf{chemo}   & \textbf{cancer}   & \textbf{really} & tysabri & \textbf{abilify} & \textbf{meds} \\
left            & xarelto          & \textbf{weeks}     & \textbf{good}      & \textbf{know}      & \textbf{good}            & \textbf{time}   & \textbf{years}   \\
right           & \textbf{effects} & \textbf{herceptin} & \textbf{time}      & use                & remicade      & vyvanse         & topamax          \\
\textbf{high}   & entresto         & cycle              & \textbf{effects}   & cream              & \textbf{weeks}           & \textbf{day}    & \textbf{day}     \\
\textbf{years}  & \textbf{time}    & \textbf{treatment} & hope               & \textbf{time}      & \textbf{know}     & latuda          & \textbf{effects} \\
\textbf{time}   & \textbf{doctor}  & \textbf{time}    & \textbf{know}     & \textbf{years}  & stelara & \textbf{zyprexa} & \textbf{feel} \\
\textbf{feel}   & \textbf{said}    & perjeta            & \textbf{avastin}   & flare              & \textbf{effects}            & \textbf{help}   & \textbf{good}    \\
normal          & \textbf{know}    & \textbf{know}      & faslodex           & \textbf{feel}      & \textbf{months}   & \textbf{meds}   & \textbf{know}    \\
\textbf{doctor} & \textbf{taking}  & \textbf{avastin}   & \textbf{feel}      & bad                & \textbf{started}    & \textbf{years}  & mg               \\
pressure        & coumadin         & days               & \textbf{weeks}     & \textbf{help}      & cosentyx   & champix         & \textbf{work}    \\
\textbf{help}   & warfarin         & \textbf{months}    & phesgo             & body               & enbrel           & anxiety         & \textbf{weight}  \\
\textbf{ago}    & \textbf{ago}     & e2                 & oncologist         & dry                & \textbf{psoriasis}             & \textbf{know}   & symptoms         \\
\textbf{pain}   & \textbf{think}   & \textbf{good}      & \textbf{said}      & itchy              & infusion & \textbf{work}   & sleep            \\
hospital        & \textbf{pain}    & \textbf{taking}    & \textbf{months}    & scalp              & \textbf{pain}           & efexor          & tried            \\
\textbf{said}   & \textbf{high}    & \textbf{started} & \textbf{arimidex} & worse           & drug & \textbf{effects} & risperdal     \\
\textbf{think}  & \textbf{meds}    & \textbf{effects}   & scan               & hands              & biologic               & \textbf{weight} & migraines        \\
\bottomrule
\end{tabular}
\label{tab:tfidf_ranked}
\end{table}

These insights underscore the varied linguistic landscapes within patient narratives across therapeutic domains and platforms, highlighting the importance of nuanced analysis in understanding patient discourse.

\subsection*{Text classification}

Building upon the insights gathered from our linguistic analyses, we explore the application of NLP models to identify patient voices within our datasets.
This exploration involves a series of experiments across varied therapeutic domains and data sources.
Each experiment is performed using two different classifier model architectures, a baseline CNN model from spaCy~\cite{honnibal_2016} and a Transformer model ~\cite{liu2019roberta}.
We report the performance of each classifier model trained on a data source and therapeutic domain specific dataset, or on a combination of these datasets.

\subsubsection*{Experiment 1: Data source and therapeutic domain specific classifiers}

\begin{table}[t]
\centering
\caption{Data source and therapeutic domain specific classifiers precision, recall and F1 score evaluation scores, each classifier evaluated on their own test datasets.}
\begin{tabular}{llcccccc}
\toprule
\multicolumn{2}{l}{\multirow{2}{*}{\textbf{Classifier}}} & \multicolumn{3}{c}{\textbf{CNN}}                          & \multicolumn{3}{c}{\textbf{Transformer}}                  \\
\multicolumn{2}{l}{}                            & \textbf{Precision}       & \textbf{Recall}         & \textbf{F1}       & \textbf{Precision}       & \textbf{Recall}       & \textbf{F1}      \\
\midrule
\multirow{2}{*}{Cardiovascular} & Reddit & 0.742 & \textbf{0.915} & 0.819          & \textbf{0.839} & 0.891 & \textbf{0.865} \\
                               & SocialGist     & \textbf{0.635} & 0.946          & \textbf{0.760} & 0.513          & \textbf{0.983} & 0.674          \\
\midrule
\multirow{2}{*}{Oncology}      & Reddit         & 0.945          & \textbf{0.986} & \textbf{0.980} & \textbf{1.0}   & 0.808          & 0.894          \\
                               & SocialGist     & 0.947          & \textbf{0.971} & 0.959          & \textbf{0.970} & 0.953          & \textbf{0.961} \\
\midrule
\multirow{2}{*}{Immunology}     & Reddit & 0.947 & \textbf{0.910} & \textbf{0.928} & \textbf{0.972} & 0.881 & 0.924          \\
                               & SocialGist     & 0.959          & 0.977          & 0.968          & \textbf{0.977} & \textbf{0.981} & \textbf{0.979} \\
\midrule
\multirow{2}{*}{Neurology}     & Reddit         & 0.995          & 0.973          & 0.984          & \textbf{1.0}   & \textbf{1.0}   & \textbf{1.0}   \\
                               & SocialGist     & 0.693          & 0.972          & 0.809          & \textbf{0.894} & \textbf{1.0}   & \textbf{0.944} \\
\bottomrule
\end{tabular}
\label{tab:tads_eval}
\end{table}

Our initial experiment aims to understand the nuanced language differences among patient groups, by training and evaluating separate classifiers for each data source and therapeutic domain.
This approach allows us to gauge the effectiveness of both model architectures in a specified setting.
As detailed in~\autoref{tab:tads_eval}, we present the precision, recall, and F1 scores for each model.
Notably, the F1 scores for both classifiers span from 0.928 to 1.0 across most data subsets.
An exception is observed in the cardiovascular classifiers for Reddit (\texttt{cardiovascular\_r}) and SocialGist (\texttt{cardiovascular\_sg}), where F1 scores dipped to 0.865 and 0.760, respectively.
In this setting, the Transformer model consistently outperforms the CNN in precision and F1 scores, although both architectures demonstrated equivalent recall performance.

\subsubsection*{Experiment 2: Combined therapeutic domain, data source and all data classifiers}

\begin{table}[t]
\centering
\caption{Data source specific, therapeutic domain specific and all data classifiers precision, recall and F1 score evaluation scores, each classifier evaluated on their own test datasets.}
\begin{tabular}{lcccccc}
\toprule
\multirow{2}{*}{\textbf{Classifier}} & \multicolumn{3}{c}{\textbf{CNN}}                          & \multicolumn{3}{c}{\textbf{Transformer}}                  \\
                            & \textbf{Precision}      & \textbf{Recall}         & \textbf{F1}       & \textbf{Precision}      & \textbf{Recall}         & \textbf{F1}       \\
\midrule
Cardiovascular              & 0.896          & \textbf{0.962} & 0.928          & \textbf{0.921} & 0.960          & \textbf{0.940} \\
Oncology                    & \textbf{0.936} & 0.961          & \textbf{0.948} & 0.857          & \textbf{0.991} & 0.919          \\
Immunology                  & \textbf{0.865} & 0.934          & 0.898          & 0.845          & \textbf{0.989} & \textbf{0.911} \\
Neurology                   & 0.920          & \textbf{0.947} & \textbf{0.933} & \textbf{0.959} & 0.878          & 0.917          \\
\midrule
Reddit COIN                 & \textbf{0.892} & 0.919          & 0.905          & \textbf{0.892} & \textbf{0.971} & \textbf{0.930} \\
SocialGist COIN             & \textbf{0.940} & 0.884          & \textbf{0.911} & 0.676          & \textbf{1.0}   & 0.806          \\
\midrule
All                         & 0.866          & \textbf{0.968} & 0.915          & \textbf{0.942} & 0.936          & \textbf{0.939} \\
\bottomrule
\end{tabular}
\label{tab:combined_eval}
\end{table}

In our second experiment, we aimed to examine the model while leveraging the lexical similarities identified in the TF-IDF analysis.
To this end, we combine datasets within the same therapeutic domain (e.g. merging \texttt{cardiovascular\_r} and \texttt{cardiovascular\_sg} into a single \texttt{cardiovascular} dataset) and from the same data source (e.g. aggregating \texttt{cardiovascular\_r}, \texttt{oncology\_r}, \texttt{immunology\_r}, and \texttt{neurology\_r} into \texttt{reddit\_coin}).
Additionally, an \texttt{all} dataset was created to encompass all data collected.
This approach was motivated by the hypothesis that datasets with similar vocabularies, as revealed through TF-IDF analysis, could benefit from combined training, potentially yielding classifiers with improved generalizability.
Consequently, our focus was on comparing the effectiveness of both CNN and Transformer model architectures across these aggregated datasets.

The results of this experiment, as detailed in~\autoref{tab:combined_eval}, underscore the performance variations between our models.
Notably, the transformer classifiers demonstrate a marginal advantage over CNN models in recall and F1 scores, indicating their robustness in more generalized settings.
The highest F1 scores observed ranged from 0.911 to 0.948, with notable performance by the \texttt{immunology} and \texttt{socialgist\_coin} classifiers.
The low cosine similarity score between the two immunology datasets as shown in the TF-IDF analysis may explain the relatively low F1 score achieved by the \texttt{immunology} classifier.
Despite these differences, precision scores were consistently comparable across both architectures.

\subsubsection*{Experiment 3: All classifiers comparison on each therapeutic domain and data source specific dataset.}

In our third experiment, we assess the performance of classifiers trained on specific combinations of therapeutic domains and data sources.
This evaluation aimed to discern the optimal classifier configuration for each unique dataset scenario.
One example is the evaluation of the Reddit cardiovascular dataset (\texttt{cardiovascular\_r}), which was tested against classifiers trained on \texttt{cardiovascular\_r}, the aggregated \texttt{cardiovascular} dataset, the \texttt{reddit\_coin} dataset representing a collection from the same data source, and the \texttt{all} dataset encompassing the entirety of our collected data.

\autoref{tab:all_eval} details the precision, recall, and F1 scores of these classifiers.
Across the datasets, the highest F1 scores varied notably, ranging from 0.977 to a perfect 1.0.
An exception was observed within the cardiovascular datasets from Reddit and SocialGist, where F1 scores were marginally lower, at 0.865 and 0.863, respectively.
Remarkably, the transformer-based classifiers consistently outperformed the CNN models in nearly all metrics, signifying their capability to handle the complexity of patient language across diverse medical discussions.
The only exception to this trend was noted in the recall metric for the Reddit neurology dataset, where a CNN model narrowly outperformed the transformer counterpart.

\begin{table}[htbp]
\centering
\caption{All experiment classifiers, evaluated on therapeutic domain and data source-specific test datasets. For each test dataset, the precision, recall and F1 score are compared between the all data classifiers, the data source specific classifier, the therapeutic domain classifier, and the therapeutic domain and data source specific classifier. A \textbf{bold cell} indicates the highest performance in a test dataset.}
\begin{tabular}{lllcccccc}
\toprule
\multicolumn{2}{l}{\textbf{Test}}             & \multirow{2}{*}{\textbf{Classifier}} & \multicolumn{3}{c}{\textbf{CNN}} & \multicolumn{3}{c}{\textbf{Transformer}}                \\
\multicolumn{2}{l}{\textbf{Dataset}}           &                           & \textbf{Precision}     & \textbf{Recall}            & \textbf{F1}    & \textbf{Precision}     & \textbf{Recall}            & \textbf{F1}             \\
\midrule
\multirow{8}{*}{\rotatebox[origin=c]{90}{Cardiovascular}} & \multirow{4}{*}{\rotatebox[origin=c]{90}{Reddit}}     & Cardiovascular Reddit       & 0.742  & 0.915  & 0.819 & \textbf{0.839} & 0.891        & \textbf{0.865} \\
 &                             & Cardiovascular            & 0.738 & 0.938        & 0.826 & 0.726          & \textbf{0.977} & 0.833          \\
 &                             & Reddit COIN               & 0.751 & 0.922        & 0.828 & 0.740          & 0.969          & 0.839          \\
 &                             & All                       & 0.719 & 0.961        & 0.823 & 0.790          & 0.950          & 0.863          \\ \cline{2-9} 
 & \multirow{4}{*}{\rotatebox[origin=c]{90}{SocialGist}} & Cardiovascular SocialGist & 0.635 & 0.946        & 0.760 & 0.513          & 0.983          & 0.674          \\
 &                             & Cardiovascular            & 0.738 & 0.938        & 0.826 & 0.726          & 0.977          & 0.833          \\
 &                             & SocialGist COIN           & 0.679 & 0.969        & 0.799 & 0.516          & \textbf{1.0}   & 0.681          \\
 &                             & All                       & 0.719 & 0.961        & 0.823 & \textbf{0.790} & 0.950          & \textbf{0.863} \\
\midrule
\multirow{8}{*}{\rotatebox[origin=c]{90}{Oncology}}       & \multirow{4}{*}{\rotatebox[origin=c]{90}{Reddit}}     & Oncology Reddit             & 0.945  & 0.986  & 0.980 & \textbf{1.0}   & 0.808        & 0.894          \\
 &                             & Oncology                  & 0.961 & 0.989        & 0.975 & 0.962          & 0.996          & 0.979          \\
 &                             & Reddit COIN               & 0.967 & 0.964        & 0.966 & 0.982          & \textbf{1.0}   & 0.991          \\
 &                             & All                       & 0.955 & 0.996        & 0.975 & \textbf{1.0}   & 0.986          & \textbf{0.993} \\ \cline{2-9} 
 & \multirow{4}{*}{\rotatebox[origin=c]{90}{SocialGist}} & Oncology SocialGist       & 0.947 & 0.971        & 0.959 & 0.970          & 0.953          & 0.961          \\
 &                             & Oncology                  & 0.947 & 0.975        & 0.961 & 0.932          & 0.996          & 0.963          \\
 &                             & SocialGist COIN           & 0.948 & 0.993        & 0.970 & 0.923          & \textbf{1.0}   & 0.960          \\
 &                             & All                       & 0.938 & 0.996        & 0.967 & \textbf{0.971} & 0.989          & \textbf{0.980} \\
\midrule
\multirow{8}{*}{\rotatebox[origin=c]{90}{Immunology}}     & \multirow{4}{*}{\rotatebox[origin=c]{90}{Reddit}}     & Immunology Reddit           & 0.947  & 0.910  & 0.928 & \textbf{0.972} & 0.881        & 0.924          \\
 &                             & Immunology                & 0.963 & 0.939        & 0.951 & 0.949          & \textbf{1.0}   & 0.974          \\
 &                             & Reddit COIN               & 0.957 & 0.975        & 0.966 & 0.965          & 0.986          & 0.975          \\
 &                             & All                       & 0.963 & 0.931        & 0.947 & 0.968          & 0.986          & \textbf{0.977} \\ \cline{2-9} 
 & \multirow{4}{*}{\rotatebox[origin=c]{90}{SocialGist}} & Immunology SocialGist     & 0.959 & 0.977        & 0.968 & 0.977          & 0.981          & 0.979          \\
 &                             & Immunology                & 0.943 & 0.950        & 0.947 & 0.956          & 0.996          & 0.976          \\
 &                             & SocialGist COIN           & 0.952 & 0.989        & 0.970 & 0.900          & \textbf{1.0}   & 0.947          \\
 &                             & All                       & 0.959 & 0.985        & 0.972 & \textbf{0.981} & 0.981          & \textbf{0.981} \\
\midrule
\multirow{8}{*}{\rotatebox[origin=c]{90}{Neurology}}      & \multirow{4}{*}{\rotatebox[origin=c]{90}{Reddit}}     & Neurology Reddit            & 0.995  & 0.973  & 0.984 & \textbf{1.0}   & \textbf{1.0} & \textbf{1.0}   \\
 &                             & Neurology                 & 0.995 & 0.973        & 0.984 & \textbf{1.0}   & 0.979          & 0.989          \\
 &                             & Reddit COIN               & 0.979 & 0.995        & 0.987 & 0.995          & 0.995          & 0.995          \\
 &                             & All                       & 0.954 & \textbf{1.0} & 0.977 & 0.995          & 0.995          & 0.995          \\ \cline{2-9} 
                                & \multirow{4}{*}{\rotatebox[origin=c]{90}{SocialGist}} & Neurology SocialGist        & 0.693  & 0.972  & 0.809 & 0.894          & \textbf{1.0} & 0.944          \\
 &                             & Neurology                 & 0.814 & 0.944        & 0.875 & \textbf{1.0}   & 0.986          & \textbf{0.993} \\
 &                             & SocialGist COIN           & 0.781 & 0.993        & 0.875 & 0.571          & \textbf{1.0}   & 0.727          \\
 &                             & All                       & 0.753 & 0.993        & 0.856 & 0.986          & \textbf{1.0}   & \textbf{0.993} \\
\bottomrule
\end{tabular}
\label{tab:all_eval}
\end{table}

In conclusion, the results of our classifier modeling experiments reflect the observations made in the linguistic and statistical text analysis.
The TF-IDF similarity analysis showed similarity between patients' language within the same therapeutic domain across both data sources.
Observing the achieved F1 scores in~\autoref{tab:all_eval}, the classifiers trained on combined datasets, outperform the classifiers trained on therapeutic domain and data source specific datasets, with the exception of the Reddit cardiovascular and Reddit neurology classifiers which slightly outperformed classifiers trained on combined datasets.
These results confirm the efficacy of combining datasets with linguistic similarities as identified by TF-IDF.

Our analyses further reveal two findings.
First, the transformer model architecture consistently outperforms the CNN model architecture, in all three metrics of precision, recall and F1 scores.
This pattern is observable across all conducted experiments, suggesting a considerable advantage of the transformer model’s pre-training.
Furthermore, the results indicate that transformer classifiers achieve higher F1 scores when trained on combined datasets, in comparison to those trained on more specific datasets.
The \texttt{all} transformer classifier, in particular, demonstrates this trend by achieving the highest F1 scores across comparisons, albeit with marginal exceptions noted in specific Reddit datasets related to cardiovascular and neurology topics.

On the other hand, the CNN classifier model achieved the highest F1 scores predominantly with data source-specific classifiers, followed by the therapeutic domain-specific classifiers.
This trend suggests that the CNN model may be more effective when it is trained on datasets that are limited in scope and highly specialized.
Overall, we observed that larger and more complex transformer models, trained on more data, are the best-performing classifiers in classifying patients' posts collected from social media and message boards.

\begin{figure}[ht]
    \centering
    \includegraphics[width=\textwidth]{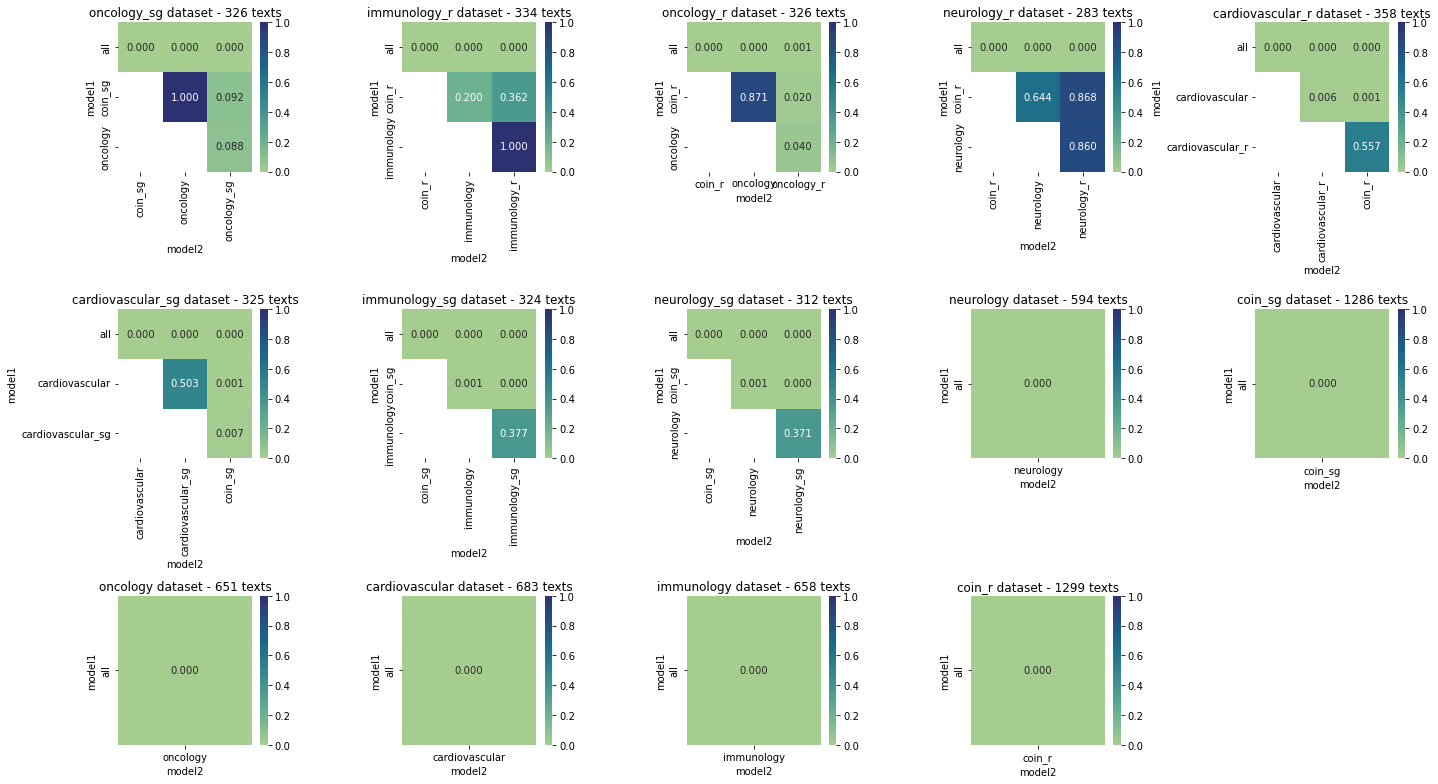}
    \caption{Heatmaps of McNemar's test p-values for pairwise comparisons of classifiers. Each heatmap corresponds to a specific test dataset. P-values greater than 0.05 suggest no significant difference in classifier performance.}
    \label{fig:mcnemar_results}
\end{figure}

To further validate our findings, we conducted McNemar's test on all pairwise comparisons of classifier predictions across each test dataset. The p-values from these tests are visualized in Figure~\ref{fig:mcnemar_results}. Any p-value greater than 0.05 indicates that the performance difference between the two classifiers is not statistically significant and likely due to random variation.

Out of 54 comparisons, 14 were not statistically significant, all occurring in the smallest datasets. This supports the reviewers' concern that small test datasets might not provide sufficient data to detect true performance differences.



\section*{Conclusions}

In this study, we explored the identification of patient voices within posts collected from social media and message boards across four distinct therapeutic domains (i.e. cardiovascular, oncology, immunology, and neurology).
We observed that each therapeutic domain can be characterized by unique linguistic features.
Through qualitative linguistic and statistical text similarity analyses, we identified specific ways patients communicate their experiences.
This analytical approach not only enabled the identification of lexical similarities across datasets but also informed the strategic aggregation of datasets for the training of NLP models.
We noticed that patients across domains, with the exception of immunology, demonstrated similar linguistic patterns across different data sources.


Our experiments highlight the effectiveness of merging linguistically similar datasets.
By creating larger, more robust training sets, we were able to enhance the performance of our classifiers.
In particular, we found that classifier models trained on such aggregated datasets consistently outperformed those trained on narrower, domain-specific datasets.
Moreover, our comparative analysis of classifier architectures revealed a clear advantage of pre-trained transformer models over CNN models for this classification task.

Conclusively, our research validates the premise that a detailed understanding of linguistic characteristics across different datasets and therapeutic domains can improve patient voice classification accuracy.
This is particularly evident when datasets sharing linguistic similarities are combined to form more expansive training corpora, further supported by the McNemar test results showing statistical significance in these larger datasets.
This approach not only leverages the inherent linguistic nuances within patient discourse but also capitalizes on the advanced capabilities of transformer models, setting a new standard in the classification of patient voices within the evolving landscape of digital health narratives.

\section*{Data availability}

We provide the list of subreddits and search terms, which we used to collect the data for this research and development project, in the supplementary material. The annotation labels and examples are also described in this paper. The third-party tools (classifiers and annotation tool) used for this work are freely available and details on the classifier set-up and model parameters are provided in this paper. For more information about this project and the data please contact Elizabeth A.L. Fairley.

\bibliography{bibliography}



\section*{Acknowledgements}


We would like to express our gratitude to the invaluable Talking Medicines Limited annotators for their hard work in creating the data needed for model training and validation. We would also like to thank Ellen Halliday, for her consultation in all compliance matters concerning this paper's experiments, as well as the Talking Medicines Founders Jo-Anne Halliday and Scott F. Crae for their support of this project.

\section*{Author contributions}



G.L. and R.E.O. contributed equally to the writing of the original draft and the review and editing of the manuscript. R.E.O. was responsible for data curation, including managing data annotation used for training, validation, evaluation, and inter-annotator agreement calculations, and performed the formal analysis on the linguistic aspects of the data. G.L. conducted the formal analysis on the statistical text similarity, inter-annotator agreement calculations, and modeling experiments. V.P. contributed to the acquisition and processing of the experiment data and the development of annotation tools. G.L.B. provided software and engineering support for the infrastructure to run the experiments. A.P.G contributed to the review and editing of the manuscript. B.A. contributed to the conceptualisation of the project, and participated in the review and editing of the manuscript. E.A.L.F. contributed to the conceptualisation and overall direction of the project, provided supervision and participated in the review and editing of the manuscript. All authors approved the submitted version and agreed to be accountable for their contributions.

\section*{Funding}

This work was funded by Talking Medicines Limited.

\section*{Additional information}

\subsection*{Competing interests}

The authors declare that they are employees of Talking Medicines Limited, which funded this research. The company may have a financial interest in the results of this study. However, the authors maintain that the research was conducted objectively and without bias.

\end{document}